\documentclass[11pt,a4paper]{article}
\usepackage[hyperref]{acl2018}
\usepackage{times}
\usepackage{latexsym}
\usepackage{graphicx}
\usepackage{url}
\usepackage{multirow}
\usepackage{subfigure}
\usepackage{amsmath}
\usepackage{xcolor}
\usepackage{bm}

\aclfinalcopy 

\title{Sentence-State LSTM for Text Representation}

\author{Yue Zhang$^1$, Qi Liu$^1$ \and Linfeng Song$^2$ \\
  $^1$Singapore University of Technology and Design \\
  $^2$Department of Computer Science, University of Rochester \\
  {\tt \{yue\_zhang, qi\_liu\}@sutd.edu.sg, lsong10@cs.rochester.edu} \\ }

\date{}

\begin{document}
\maketitle
\begin{abstract}
Bi-directional LSTMs are a powerful tool for text representation. 
On the other hand, they have been shown to suffer various limitations due to their sequential nature. 
We investigate an alternative LSTM structure for encoding text, which consists of a parallel state for each word. Recurrent steps are used to perform local and global information exchange between words simultaneously, rather than incremental reading of a sequence of words. 
Results on various classification and sequence labelling benchmarks show that the proposed model has strong representation power, giving highly competitive performances compared to stacked BiLSTM models with similar parameter numbers.
\end{abstract}

\section{Introduction}
Neural models have become the dominant approach in the NLP literature.
Compared to hand-crafted indicator features, neural sentence representations are less sparse, and more flexible in encoding intricate syntactic and semantic information. 
Among various neural networks for encoding sentences, bi-directional LSTMs (BiLSTM) \cite{hochreiter1997long} have been a dominant method, giving state-of-the-art results in language modelling \cite{sundermeyer2012lstm}, machine translation \cite{bahdanau2014neural}, syntactic parsing \cite{dozat2016deep} and question answering \cite{tan2015lstm}.

Despite their success, BiLSTMs have been shown to suffer several limitations. 
For example, their inherently sequential nature endows computation non-parallel within the same sentence \cite{vaswani2017attention}, which can lead to a computational bottleneck, hindering their use in the industry.
In addition, local ngrams, which have been shown a highly useful source of contextual information for NLP, are not explicitly modelled  \cite{wang2016combination}. 
Finally, sequential information flow leads to relatively weaker power in capturing long-range dependencies, which results in lower performance in encoding longer sentences \cite{koehn2017six}.

\begin{figure}[t]
\centering	
\includegraphics[width=0.45\textwidth]{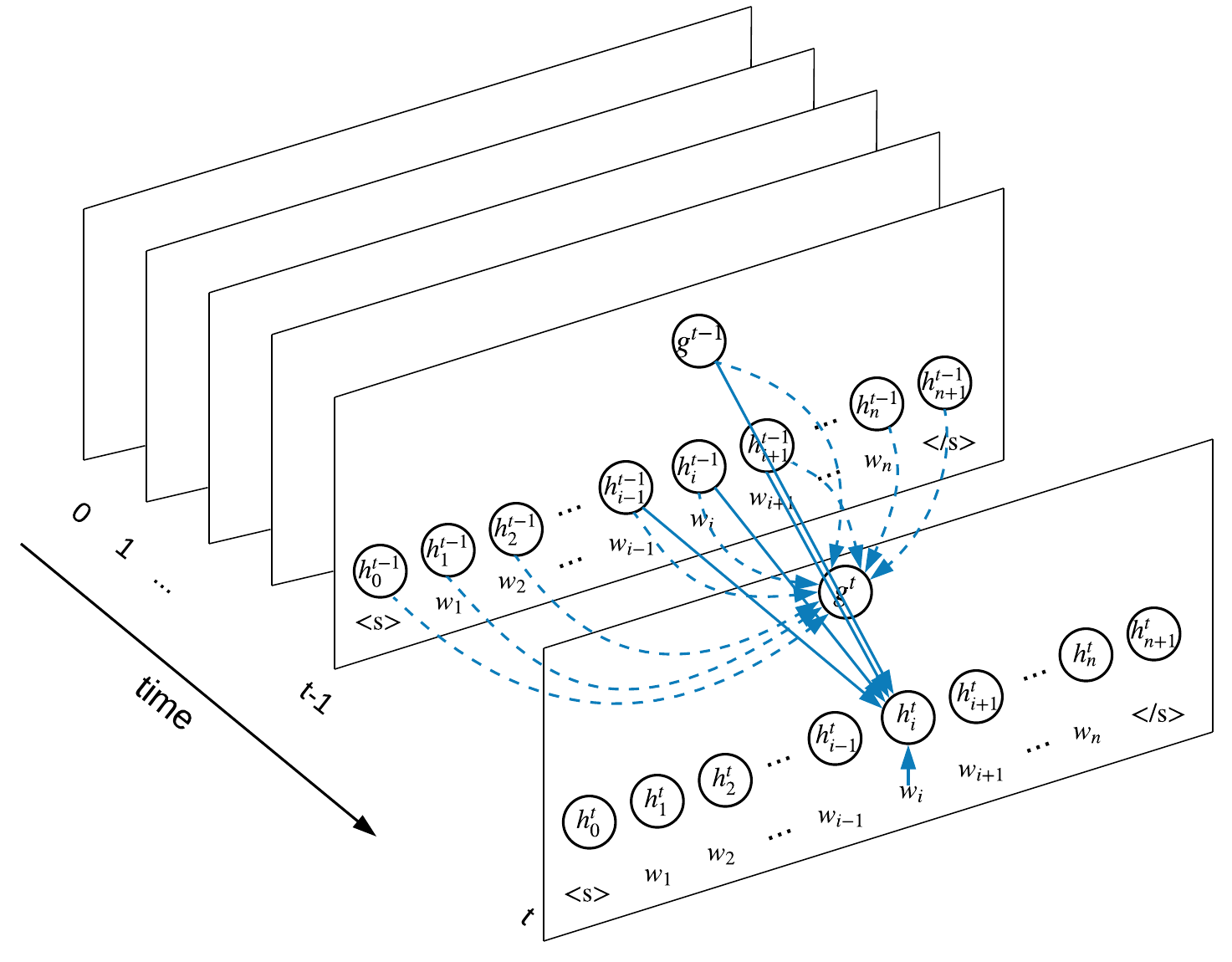}
\caption{Sentence-State LSTM}
\label{fig:mlstm}
\end{figure}

We investigate an alternative recurrent neural network structure for addressing these issues.
As shown in Figure \ref{fig:mlstm}, the main idea is to model the hidden states of all words simultaneously at each recurrent step, rather than one word at a time. 
In particular, we view the whole sentence as a single state, which consists of sub-states for individual words and an overall sentence-level state. 
To capture local and non-local contexts, states are updated recurrently by exchanging information between each other. 
Consequently, we refer to our model as sentence-state LSTM, or S-LSTM in short. 
Empirically, S-LSTM can give effective sentence encoding after 3 -- 6 recurrent steps. 
In contrast, the number of recurrent steps necessary for BiLSTM scales with the size of the sentence.

At each recurrent step, information exchange is conducted between consecutive words in the sentence, and between the sentence-level state and each word. 
In particular, each word receives information from its predecessor and successor simultaneously. 
From an initial state without information exchange, each word-level state can obtain 3-gram, 5-gram and 7-gram information after 1, 2 and 3 recurrent steps, respectively. 
Being connected with every word, the sentence-level state vector serves to exchange non-local information with each word. 
In addition, it can also be used as a global sentence-level representation for classification tasks.

Results on both classification and sequence labelling show that S-LSTM gives better accuracies compared to BiLSTM using the same number of parameters, while being faster. We release our code and models at \url{https://github.com/leuchine/S-LSTM}, which include all baselines and the final model.

\section{Related Work}

LSTM \cite{graves2005framewise} showed its early potentials in NLP when a neural machine translation system that leverages LSTM source encoding gave highly competitive results compared to the best SMT models \cite{bahdanau2014neural}. 
LSTM encoders have since been explored for other tasks, including syntactic parsing \cite{dyer2015transition}, text classification \cite{yang2016hierarchical} and machine reading \cite{hermann2015teaching}. 
Bi-directional extensions have become a standard configuration for achieving state-of-the-art accuracies among various tasks \cite{wen2015semantically,ma2016end,dozat2016deep}. 
S-LSTMs are similar to BiLSTMs in their recurrent bi-directional message flow between words, but different in the design of state transition.

CNNs \cite{krizhevsky2012imagenet} also allow better parallelisation compared to LSTMs for sentence encoding \cite{kim2014convolutional}, thanks to parallelism among convolution filters. 
On the other hand, convolution features embody only fix-sized local n-gram information, whereas sentence-level feature aggregation via pooling can lead to loss of information \cite{sabour2017dynamic}.
In contrast, S-LSTM uses a global sentence-level node to assemble and back-distribute local information in the recurrent state transition process, suffering less information loss compared to pooling.

Attention \cite{bahdanau2014neural} has recently been explored as a standalone method for sentence encoding, giving competitive results compared to Bi-LSTM encoders for neural machine translation \cite{vaswani2017attention}. 
The attention mechanism allows parallelisation, and can play a similar role to the sentence-level state in S-LSTMs, which uses neural gates to integrate word-level information compared to hierarchical attention.
S-LSTM further allows local communication between neighbouring words.

Hierarchical stacking of CNN layers \cite{lecun1995convolutional,kalchbrenner2014convolutional,papandreou2015weakly,dauphin2016language} allows better interaction between non-local components in a sentence via incremental levels of abstraction.
S-LSTM is similar to hierarchical attention and stacked CNN in this respect, incrementally refining sentence representations.
However, S-LSTM models hierarchical encoding of sentence structure as a \emph{recurrent} state transition process.
In nature, our work belongs to the family of LSTM sentence representations.

S-LSTM is inspired by message passing over graphs \cite{murphy1999loopy,scarselli2009graph}.
Graph-structure neural models have been used for computer program verification \cite{li2015gated} and image object detection \cite{liang2016semantic}. 
The closest previous work in NLP includes the use of convolutional neural networks \cite{bastings-EtAl:2017:EMNLP2017,marcheggiani-titov:2017:EMNLP2017} and DAG LSTMs \cite{TACL1028} for modelling syntactic structures.
Compared to our work, their motivations and network structures are highly different.
In particular, the DAG LSTM of \newcite{TACL1028} is a natural extension of tree LSTM \cite{tai-socher-manning:2015:ACL-IJCNLP}, and is sequential rather than parallel in nature.
To our knowledge, we are the first to investigate a graph RNN for encoding sentences, proposing parallel graph states for integrating word-level and sentence-level information.
In this perspective, our contribution is similar to that of \newcite{kim2014convolutional} and \newcite{bahdanau2014neural} in introducing a neural representation to the NLP literature.

\section{Model}
Given a sentence $\boldsymbol{s}=w_1,\, w_2, \, \dots , \, w_n$, where $w_i$ represents the $i$th word and $n$ is the sentence length, our goal is to find a neural representation of $\boldsymbol{s}$, which consists of a hidden vector $\boldsymbol{h}_i$ for each input word $w_i$, and a global sentence-level hidden vector $\boldsymbol{g}$. 
Here $\boldsymbol{h}_i$ represents syntactic and semantic features for $w_i$ under the sentential context, while $\boldsymbol{g}$ represents features for the whole sentence. 
Following previous work, we additionally add $\langle s \rangle$ and $\langle /s \rangle$ to the two ends of the sentence as $w_0$ and $w_{n+1}$, respectively.


\subsection{Baseline BiLSTM}
\label{sec:bilstm}


The baseline BiLSTM model consists of two LSTM components, which process the input in the forward left-to-right and the backward right-to-left directions, respectively. 
In each direction, the reading of input words is modelled as a recurrent process with a single hidden state. 
Given an initial value, the state changes its value recurrently, each time consuming an incoming word.

Take the forward LSTM component for example. Denoting the initial state as $\boldsymbol{\overrightarrow{h}}^0$, which is a model parameter, the recurrent state transition step for calculating $\boldsymbol{\overrightarrow{h}}^1,\dots,\boldsymbol{\overrightarrow{h}}^{n+1}$ is defined as follows \cite{graves2005framewise}:
\begin{equation} \label{eq:gate_bilstm}
\begin{split}
\boldsymbol{\hat{i}}^t &= \sigma(\boldsymbol{W}_i \boldsymbol{x}_t + \boldsymbol{U}_i \boldsymbol{\overrightarrow{h}}^{t-1} + \boldsymbol{b}_i) \\
\boldsymbol{\hat{f}}^t &= \sigma(\boldsymbol{W}_f \boldsymbol{x}_t + \boldsymbol{U}_f \boldsymbol{\overrightarrow{h}}^{t-1} + \boldsymbol{b}_f) \\
\boldsymbol{o^t} &= \sigma(\boldsymbol{W}_o \boldsymbol{x}_t + \boldsymbol{U}_o \boldsymbol{\overrightarrow{h}}^{t-1} + \boldsymbol{b}_o) \\
\boldsymbol{u^t} &= {\it tanh}(\boldsymbol{W}_u \boldsymbol{x}_t + \boldsymbol{U}_u \boldsymbol{\overrightarrow{h}}^{t-1} + \boldsymbol{b}_u) \\
\boldsymbol{i}^t&, \boldsymbol{f}^t = {\it softmax}(\boldsymbol{\hat{i}}^t,\boldsymbol{\hat{f}}^t) \\
\boldsymbol{c}^t &= \boldsymbol{c}^{t-1} \odot \boldsymbol{f}^t + \boldsymbol{u}^t \odot \boldsymbol{i}^t \\
\boldsymbol{\overrightarrow{h}}&^t = \boldsymbol{o}^t \odot {\it tanh}(\boldsymbol{c}^t)
\end{split}
\end{equation}
where $\boldsymbol{x}_t$ denotes the word representation of $w_t$; $\boldsymbol{i}^t$, $\boldsymbol{o}^t$, $\boldsymbol{f}^t$ and $\boldsymbol{u}^t$ represent the values of an input gate, an output gate, a forget gate and an actual input at time step $t$, respectively, which controls the information flow for a recurrent cell $\boldsymbol{\overrightarrow{c}}^t$ and the state vector $\boldsymbol{\overrightarrow{h}}^t$; $\boldsymbol{W}_x$, $\boldsymbol{U}_x$ and $\boldsymbol{b}_x$ ($x \in \{i,o,f,u\}$) are model parameters.
$\sigma$ is the sigmoid function.

The backward LSTM component follows the same recurrent state transition process as described in Eq \ref{eq:gate_bilstm}. 
Starting from an initial state $\boldsymbol{h}^{n+1}$, which is a model parameter, it reads the input $\boldsymbol{x}_n$, $\boldsymbol{x}_{n-1}$, $\dots$, $\boldsymbol{x}_0$, changing its value to $\boldsymbol{\overleftarrow{h}}^n$, $\boldsymbol{\overleftarrow{h}}^{n-1}$, $\dots$, $\boldsymbol{\overleftarrow{h}}^0$, respectively. 
A separate set of parameters $\boldsymbol{\hat{W}}_x$, $\boldsymbol{\hat{U}}_x$ and $\boldsymbol{\hat{b}}_x$ ($x \in \{i,o,f,u\}$) are used for the backward component.

The BiLSTM model uses the concatenated value of $\boldsymbol{\overrightarrow{h}}^t$ and $\boldsymbol{\overleftarrow{h}}^t$ as the hidden vector for $w_t$:
\[
\boldsymbol{h}^t = [\boldsymbol{\overrightarrow{h}}^t; \boldsymbol{\overleftarrow{h}}^t]
\]
A single hidden vector representation $\boldsymbol{g}$ of the whole input sentence can be obtained using the final state values of the two LSTM components:
\[
\boldsymbol{g} = [\boldsymbol{\overrightarrow{h}}^{n+1}; \boldsymbol{\overleftarrow{h}}^0]
\]


\subparagraph{Stacked BiLSTM}
Multiple layers of BiLTMs can be stacked for increased representation power, where the hidden vectors of a lower layer are used as inputs for an upper layer. 
Different model parameters are used in each stacked BiLSTM layer.

\subsection{Sentence-State LSTM}
\label{sec:m-lstm}


Formally, an S-LSTM state at time step $t$ can be denoted by:
\[
\boldsymbol{H}^t = \langle \boldsymbol{h}_0^t, \boldsymbol{h}_1^t, \dots, \boldsymbol{h}_{n+1}^t, \boldsymbol{g}^t \rangle \textrm{,}
\]
which consists of a sub state $\boldsymbol{h}_i^t$ for each word $w_i$ and a sentence-level sub state $\boldsymbol{g}^t$.

S-LSTM uses a recurrent state transition process to model information exchange between sub states, which enriches state representations incrementally.
For the initial state $\boldsymbol{H}^0$, we set $\boldsymbol{h}_i^0 = \boldsymbol{g}^0 = \boldsymbol{h}^0$, where $\boldsymbol{h}^0$ is a parameter. 
The state transition from $\boldsymbol{H}^{t-1}$ to $\boldsymbol{H}^t$ consists of sub state transitions from $\boldsymbol{h}_i^{t-1}$ to $\boldsymbol{h}_i^t$ and from $\boldsymbol{g}^{t-1}$ to $\boldsymbol{g}^t$.
We take an LSTM structure similar to the baseline BiLSTM for modelling state transition, using a recurrent cell $\boldsymbol{c}_i^t$ for each $w_i$ and a cell $\boldsymbol{c}_g^t$ for $\boldsymbol{g}$.

As shown in Figure~\ref{fig:mlstm}, the value of each $\boldsymbol{h}_i^{t}$ is computed based on the values of $\boldsymbol{x}_i$, $\boldsymbol{h}_{i-1}^{t-1}$, $\boldsymbol{h}_i^{t-1}$, $\boldsymbol{h}_{i+1}^{t-1}$ and $\boldsymbol{g}^{t-1}$, together with their corresponding cell values:
\begin{equation} \label{eq:mlstm}
\begin{split}
\boldsymbol{\xi}_i^t &= [\boldsymbol{h}_{i-1}^{t-1}, \boldsymbol{h}_i^{t-1}, \boldsymbol{h}_{i+1}^{t-1}] \\
\boldsymbol{\hat{i}}_i^t &= \sigma(\boldsymbol{W}_i \boldsymbol{\xi}_i^t + \boldsymbol{U}_i \boldsymbol{x}_{i} + \boldsymbol{V}_i \boldsymbol{g}^{t-1} + \boldsymbol{b}_i)\\
\boldsymbol{\hat{l}}_i^t &= \sigma(\boldsymbol{W}_l \boldsymbol{\xi}_i^t + \boldsymbol{U}_l \boldsymbol{x}_{i} + \boldsymbol{V}_l \boldsymbol{g}^{t-1} + \boldsymbol{b}_l)\\
\boldsymbol{\hat{r}}_i^t &= \sigma(\boldsymbol{W}_r \boldsymbol{\xi}_i^t + \boldsymbol{U}_r \boldsymbol{x}_{i} + \boldsymbol{V}_r \boldsymbol{g}^{t-1} + \boldsymbol{b}_r)\\
\boldsymbol{\hat{f}}_i^t &= \sigma(\boldsymbol{W}_f \boldsymbol{\xi}_i^t + \boldsymbol{U}_f \boldsymbol{x}_{i} + \boldsymbol{V}_f \boldsymbol{g}^{t-1} + \boldsymbol{b}_f)\\
\boldsymbol{\hat{s}}_i^t &= \sigma(\boldsymbol{W}_s \boldsymbol{\xi}_i^t + \boldsymbol{U}_s \boldsymbol{x}_{i} + \boldsymbol{V}_s \boldsymbol{g}^{t-1} + \boldsymbol{b}_s)\\
\boldsymbol{o}_i^t &= \sigma(\boldsymbol{W}_o \boldsymbol{\xi}_i^t + \boldsymbol{U}_o \boldsymbol{x}_{i} + \boldsymbol{V}_o \boldsymbol{g}^{t-1} + \boldsymbol{b}_o)\\
\boldsymbol{u}_i^t &= {\it tanh}(\boldsymbol{W}_u \boldsymbol{\xi}_i^t + \boldsymbol{U}_u \boldsymbol{x}_{i} + \boldsymbol{V}_u \boldsymbol{g}^{t-1} + \boldsymbol{b}_u)\\
\boldsymbol{i}_i^t &, \boldsymbol{l}_i^t, \boldsymbol{r}_i^t, \boldsymbol{f}_i^t, \boldsymbol{s}_i^t = {\it softmax}(\boldsymbol{\hat{i}}_i^t, \boldsymbol{\hat{l}}_i^t, \boldsymbol{\hat{r}}_i^t, \boldsymbol{\hat{f}}_i^t, \boldsymbol{\hat{s}}_i^t)\\
\boldsymbol{c}_i^t &= \boldsymbol{l}_i^t \odot \boldsymbol{c}_{i-1}^{t-1} + \boldsymbol{f}_i^t \odot \boldsymbol{c}_{i}^{t-1} + \boldsymbol{r}_i^t \odot \boldsymbol{c}_{i+1}^{t-1}\\
~~~~~~&~~~~~~~~~~~~~~~~~~~~ + \boldsymbol{s}_i^t \odot \boldsymbol{c}_g^{t-1} + \boldsymbol{i}_i^t \odot \boldsymbol{u}_i^t\\
\boldsymbol{h}_i^t &= \boldsymbol{o}_t^i \odot {\it tanh}(\boldsymbol{c}_i^t)
\end{split}
\end{equation}
where $\boldsymbol{\xi}_i^t$ is the concatenation of hidden vectors of a context window, and $\boldsymbol{l}_i^t$, $\boldsymbol{r}_i^t$, $\boldsymbol{f}_i^t$, $\boldsymbol{s}_i^t$ and $\boldsymbol{i}_i^t$ are gates that control information flow from $\boldsymbol{\xi}_i^t$ and $\boldsymbol{x}_i$ to $\boldsymbol{c}_i^t$.
In particular, $\boldsymbol{i}_i^t$ controls information from the input $\boldsymbol{x}_i$; $\boldsymbol{l}_i^t$, $\boldsymbol{r}_i^t$, $\boldsymbol{f}_i^t$ and $\boldsymbol{s}_i^t$ control information from the left context cell $\boldsymbol{c}_{i-1}^{t-1}$, the right context cell $\boldsymbol{c}_{i+1}^{t-1}$, $\boldsymbol{c}_{i}^{t-1}$ and the sentence context cell $\boldsymbol{c}_g^{t-1}$, respectively.
The values of $\boldsymbol{i}_i^t$, $\boldsymbol{l}_i^t$, $\boldsymbol{r}_i^t$, $\boldsymbol{f}_i^t$ and $\boldsymbol{s}_i^t$ are normalised such that they sum to {\bf 1}. 
$\boldsymbol{o}_i^t$ is an output gate from the cell state $\boldsymbol{c}_i^t$ to the hidden state $\boldsymbol{h}_i^t$.
$\boldsymbol{W}_x$, $\boldsymbol{U}_x$, $\boldsymbol{V}_x$ and $\boldsymbol{b}_x$ ($x \in \{i,o,l,r,f,s,u\}$) are model parameters. $\sigma$ is the sigmoid function.

The value of $\boldsymbol{g}^{t}$ is computed based on the values of $\boldsymbol{h}_i^{t-1}$ for all $i \in [0..n+1]$:
\begin{equation} \label{eq:mlstm_g}
\begin{aligned}
\boldsymbol{\bar{h}}&= {\it avg}(\boldsymbol{h}_0^{t-1}, \boldsymbol{h}_1^{t-1}, \dots, \boldsymbol{h}_{n+1}^{t-1})\\
\boldsymbol{\hat{f}}_g^{t} &= \sigma(\boldsymbol{W}_g \boldsymbol{g}^{t-1} + \boldsymbol{U}_g \boldsymbol{\bar{h}} + \boldsymbol{b}_g)\\
\boldsymbol{\hat{f}}_i^{t} &= \sigma(\boldsymbol{W}_f \boldsymbol{g}^{t-1} + \boldsymbol{U}_f \boldsymbol{h}_i^{t-1} + \boldsymbol{b}_f)\\
\boldsymbol{o}^{t} &= \sigma(\boldsymbol{W}_o \boldsymbol{g}^{t-1} + \boldsymbol{U}_o \boldsymbol{\bar{h}} + \boldsymbol{b}_o)\\
\boldsymbol{f}_0^t&, \dots, \boldsymbol{f}_{n+1}^t, \boldsymbol{f}_g^{t} = {\it softmax}(\boldsymbol{\hat{f}}_0^t, \dots, \boldsymbol{\hat{f}}_{n+1}^t, \boldsymbol{\hat{f}}_g^{t})\\
\boldsymbol{c}_g^{t} &= \boldsymbol{f}_g^{t} \odot \boldsymbol{c}_g^{t-1} + \sum_i \boldsymbol{f}_i^t \odot \boldsymbol{c}_i^{t-1}  \\
\boldsymbol{g}^t &= \boldsymbol{o}^{t} \odot {\it tanh}(\boldsymbol{c}_g^{t})\\
\end{aligned}
\end{equation}
where $\boldsymbol{f}_0^t, \dots, \boldsymbol{f}_{n+1}^t$ and $\boldsymbol{f}_g^{t}$ are gates controlling information from $\boldsymbol{c}_0^{t-1}, \dots, \boldsymbol{c}_{n+1}^{t-1}$ and $\boldsymbol{c}_g^{t-1}$, respectively, which are normalised. 
$\boldsymbol{o}^{t}$ is an output gate from the recurrent cell $\boldsymbol{c}_g^{t}$ to $\boldsymbol{g}^{t}$.
$\boldsymbol{W}_x$, $\boldsymbol{U}_x$ and $\boldsymbol{b}_x$ ($x \in \{g, f, o\}$) are model parameters.

\subparagraph{Contrast with BiLSTM}
The difference between S-LSTM and BiLSTM can be understood with respect to their recurrent states. 
While BiLSTM uses only one state in each direction to represent the subsequence from the beginning to a certain word, S-LSTM uses a structural state to represent the full sentence, which consists of a sentence-level sub state and $n+2$ word-level sub states, simultaneously.
Different from BiLSTMs, for which $\boldsymbol{h}^t$ at different time steps are used to represent $w_0, \dots, w_{n+1}$, respectively, the word-level states $\boldsymbol{h}_i^t$ and sentence-level state $\boldsymbol{g}^t$ of S-LSTMs directly correspond to the goal outputs $\boldsymbol{h}_i$ and $\boldsymbol{g}$, as introduced in the beginning of this section. 
As $t$ increases from 0, $\boldsymbol{h}_i^t$ and $\boldsymbol{g}^t$ are enriched with increasingly deeper context information.

From the perspective of information flow, BiLSTM passes information from one end of the sentence to the other. 
As a result, the number of time steps scales with the size of the input. 
In contrast, S-LSTM allows bi-directional information flow at each word simultaneously, and additionally between the sentence-level state and every word-level state. 
At each step, each $\boldsymbol{h}_i$ captures an increasing larger ngram context, while additionally communicating globally to all other $\boldsymbol{h}_j$ via $\boldsymbol{g}$.
The optimal number of recurrent steps is decided by the end-task performance, and does not necessarily scale with the sentence size. 
As a result, S-LSTM can potentially be both more efficient and more accurate compared with BiLSTMs.

{\bf Increasing window size}. 
By default S-LSTM exchanges information only between neighbouring words, which can be seen as adopting a 1-word window on each side. 
The window size can be extended to 2, 3 or more words in order to allow more communication in a state transition, expediting information exchange. 
To this end, we modify Eq~\ref{eq:mlstm}, integrating additional context words to $\boldsymbol{\xi}_i^t$, with extended gates and cells. 
For example, with a window size of 2, $\boldsymbol{\xi}_i^t = [\boldsymbol{h}_{i-2}^{t-1}, \boldsymbol{h}_{i-1}^{t-1}, \boldsymbol{h}_{i}^{t-1}, \boldsymbol{h}_{i+1}^{t-1}, \boldsymbol{h}_{i+2}^{t-1}]$.
We study the effectiveness of window size in our experiments.

{\bf Additional sentence-level nodes}. By default S-LSTM uses one sentence-level node. 
One way of enriching the parameter space is to add more sentence-level nodes, each communicating with word-level nodes in the same way as described by Eq~\ref{eq:mlstm_g}. 
In addition, different sentence-level nodes can communicate  with each other during state transition. 
When one sentence-level node is used for classification outputs, the other sentence-level node can serve as hidden memory units, or latent features. We study the effectiveness of multiple sentence-level nodes empirically.

\subsection{Task settings}
\label{sec:ext_attn}

We consider two task settings, namely classification and sequence labelling. 
For \emph{classification}, $\boldsymbol{g}$ is fed to a {\it softmax} classification layer:
\[
\boldsymbol{y} = {\it softmax}(\boldsymbol{W}_c \boldsymbol{g} + \boldsymbol{b}_c)
\]
where $\boldsymbol{y}$ is the probability distribution of output class labels and $\boldsymbol{W}_c$ and $\boldsymbol{b}_c$ are model parameters. 
For \emph{sequence labelling}, each $\boldsymbol{h}_i$ can be used as feature representation for a corresponding word $w_i$.

\subparagraph{External attention}
It has been shown that summation of hidden states using attention \cite{bahdanau2014neural,yang2016hierarchical} give better accuracies compared to using the end states of BiLSTMs. 
We study the influence of attention on both S-LSTM and BiLSTM for \emph{classification}. 
In particular, additive attention \cite{bahdanau2014neural} is applied to the hidden states of input words for both BiLSTMs and S-LSTMs calculating a weighted sum
\begin{equation*}
\boldsymbol{g} = \sum_t \alpha_t \boldsymbol{h}_t
\end{equation*}
where
\begin{equation*}
\alpha_t = \frac{\exp{\boldsymbol{u}^T \boldsymbol{\epsilon}_t}}{\sum_i \exp{\boldsymbol{u}^T\boldsymbol{\epsilon}_i}}
\end{equation*}
\begin{equation*}
\boldsymbol{\epsilon}_t= {\it tanh}(\boldsymbol{W}_\alpha \boldsymbol{h}_t+\boldsymbol{b}_\alpha)
\end{equation*}
Here $\boldsymbol{W}_\alpha$, $\boldsymbol{u}$ and $\boldsymbol{b}_\alpha$ are model parameters.

\subparagraph{External CRF}
For \emph{sequential labelling}, we use a CRF layer on top of the hidden vectors $\boldsymbol{h}_1, \boldsymbol{h}_2, \dots, \boldsymbol{h}_n$ for calculating the conditional probabilities of label sequences \cite{huang2015bidirectional,ma2016end}:
\[
P(\boldsymbol{Y}_1^n|\boldsymbol{h}, \boldsymbol{W}_s, \boldsymbol{b}_s) = \frac{\prod_{i=1}^n \psi_i(y_{i-1}, y_i, \boldsymbol{h})}{\sum_{\boldsymbol{Y}_1^{n'}} \prod_{i=1}^n \psi_i(y'_{i-1}, y'_i, \boldsymbol{h})}
\] 
\[
\psi_i(y_{i-1}, y_i, \boldsymbol{h})= {\it exp}(\boldsymbol{W}_{s}^{y_{i-1},y_i} h_i + b_{s}^{y_{i-1},y_i})
\]
where $\boldsymbol{W}_{s}^{y_{i-1},y_i}$ and $b_{s}^{y_{i-1},y_i}$ are parameters specific to two consecutive labels $y_{i-1}$ and $y_i$.

For training, standard log-likelihood loss is used with $L_2$ regularization given a set of gold-standard instances.

\section{Experiments}
We empirically compare S-LSTMs and BiLSTMs on different classification and sequence labelling tasks. 
All experiments are conducted using a GeForce GTX 1080 GPU with 8GB memory. 

\begin{table}[t]
	\centering
	\scriptsize
	\tabcolsep=0.03cm
	\begin{tabular}{|ccc|c|c|c|c|c|c|}
		\hline
		\multicolumn{3}{|c|}{\textbf{Dataset}} &\multicolumn{2}{|c|}{\textbf{Training}} &\multicolumn{2}{|c|}{\textbf{Development}}& \multicolumn{2}{|c|}{\textbf{Test}}\\
		\hline
		\multicolumn{3}{|c|}{} &\textbf{\#sent}&\textbf{\#words} &\textbf{\#sent}&\textbf{\#words}& \textbf{\#sent}&\textbf{\#words}\\
		\hline
		\multicolumn{3}{|c|}{Movie review \cite{pang2008opinion}} &8527&201137&1066&25026&1066&25260\\
		\hline
		  &  \multicolumn{2}{|c|}{Books} &1400 &297K&200&59K&400&68K\\
		  & \multicolumn{2}{|c|}{Electronics} &1398 &924K&200&184K&400&224K\\
		  & \multicolumn{2}{|c|}{DVD}&1400 &1,587K&200&317K&400&404K\\
		  & \multicolumn{2}{|c|}{Kitchen}&1400 &769K&200&153K&400&195K\\
		  & \multicolumn{2}{|c|}{Apparel}&1400 &525K&200&105K&400&128K\\
		  & \multicolumn{2}{|c|}{Camera}&1397 &1,084K&200&216K&400&260K\\
		Text  & \multicolumn{2}{|c|}{Health}&1400&742K&200&148K&400&175K\\
		Classification  & \multicolumn{2}{|c|}{Music}&1400 &1,176K&200&235K&400&276K\\
		\cite{liu2017adversarial}  & \multicolumn{2}{|c|}{Toys} &1400&792K&200&158K&400&196K\\
		  & \multicolumn{2}{|c|}{Video}& 1400 &1,311K&200&262K&400&342K\\
		  & \multicolumn{2}{|c|}{Baby}& 1300 &855K&200&171K&400&221K\\
		  & \multicolumn{2}{|c|}{Magazines}& 1370 &1,033K&200&206K&400&264K\\
		  & \multicolumn{2}{|c|}{Software}& 1315 &1,143K&200&228K&400&271K\\
		  & \multicolumn{2}{|c|}{Sports}& 1400 &833K&200&183K&400&218K\\
		  & \multicolumn{2}{|c|}{IMDB}& 1400 &2,205K&200&507K&400&475K\\
		  & \multicolumn{2}{|c|}{MR}& 1400 &196K&200&41K&400&48K\\
		\hline
		\multicolumn{3}{|c|}{POS tagging \cite{marcus1993building}}&39831&950011&1699&40068&2415&56671\\
		\hline
		\multicolumn{3}{|c|}{NER \cite{tjong2003introduction}}&14987&204567 &3466&51578&3684&46666\\
		
		\hline
	\end{tabular}
	\caption{\label{dataset_statistics}Dataset statistics}
\end{table}

\subsection{Experimental Settings}
{\bf Datasets}. We choose the movie review dataset of \newcite{pang2008opinion}, and additionally the 16 datasets of \newcite{liu2017adversarial} for classification evaluation. 
We randomly split the movie review dataset into training (80\%), development (10\%) and test (10\%) sections, and the original split of \newcite{liu2017adversarial} for the 16 classification datasets.

For sequence labelling, we choose the Penn Treebank \cite{marcus1993building} POS tagging task and the CoNLL \cite{tjong2003introduction} NER task as our benchmarks. 
For POS tagging, we follow the standard split \cite{manning2011part}, using sections 0 -- 18 for training, 19 -- 21 for development and 22 -- 24 for test. 
For NER, we follow the standard split, and use the BIOES tagging scheme \cite{ratinov2009design}. 
Statistics of the four datasets are shown in Table \ref{dataset_statistics}.

{\bf Hyperparameters}. We initialise word embeddings using GloVe \cite{pennington2014glove} 300 dimensional embeddings.\footnote{https://nlp.stanford.edu/projects/glove/}
Embeddings are fine-tuned during model training for all tasks. 
Dropout \cite{srivastava2014dropout} is applied to embedding hidden states, with a rate of 0.5. 
All models are optimised using the Adam optimizer \cite{kingma2014adam}, with an initial learning rate of 0.001 and a decay rate of 0.97. Gradients are clipped at 3 and a batch size of 10 is adopted. 
Sentences with similar lengths are batched together.
The L2 regularization parameter is set to 0.001.

\begin{table}[t]
	\centering
	\tabcolsep=0.1cm
	\begin{tabular}{|ccc|ccc|}
		\hline
		\multicolumn{3}{|c|}{\textbf{Model}}& \textbf{Time (s)} & \textbf{Acc} & \textbf{\# Param}\\ 
		\hline

		\multicolumn{3}{|c|}{+0 dummy node}&56&81.76 &7,216K\\ 
		\multicolumn{3}{|c|}{+1 dummy node}&65&82.64&8,768K\\ 
		\multicolumn{3}{|c|}{+2 dummy node}&76&82.24 &10,321K\\ 
		\hline
		\multicolumn{3}{|c|}{Hidden size 100 }&42&81.75 &4,891K\\
		\multicolumn{3}{|c|}{Hidden size 200}&54&82.04 &6,002K\\ 
		\multicolumn{3}{|c|}{Hidden size 300}&65&82.64 &8,768K\\ 
		\multicolumn{3}{|c|}{Hidden size 600}&175&81.84&17,648K\\
		\multicolumn{3}{|c|}{Hidden size 900}&235&81.66 &33,942K\\ 
		\hline
		\multicolumn{3}{|c|}{Without $\langle \textrm{s} \rangle$,  $\langle \textrm{/s} \rangle$}&63&82.36&8,768K\\ 
		\multicolumn{3}{|c|}{With $\langle \textrm{s} \rangle$,  $\langle \textrm{/s} \rangle$}&65&82.64&8,768K \\
		\hline
	\end{tabular}
	\caption{\label{tab:movie_dev}Movie review \textsc{Dev} results of S-LSTM}
\end{table}

\begin{figure}
\vspace{-0.4em}
\centering
\includegraphics[width=0.9\linewidth]{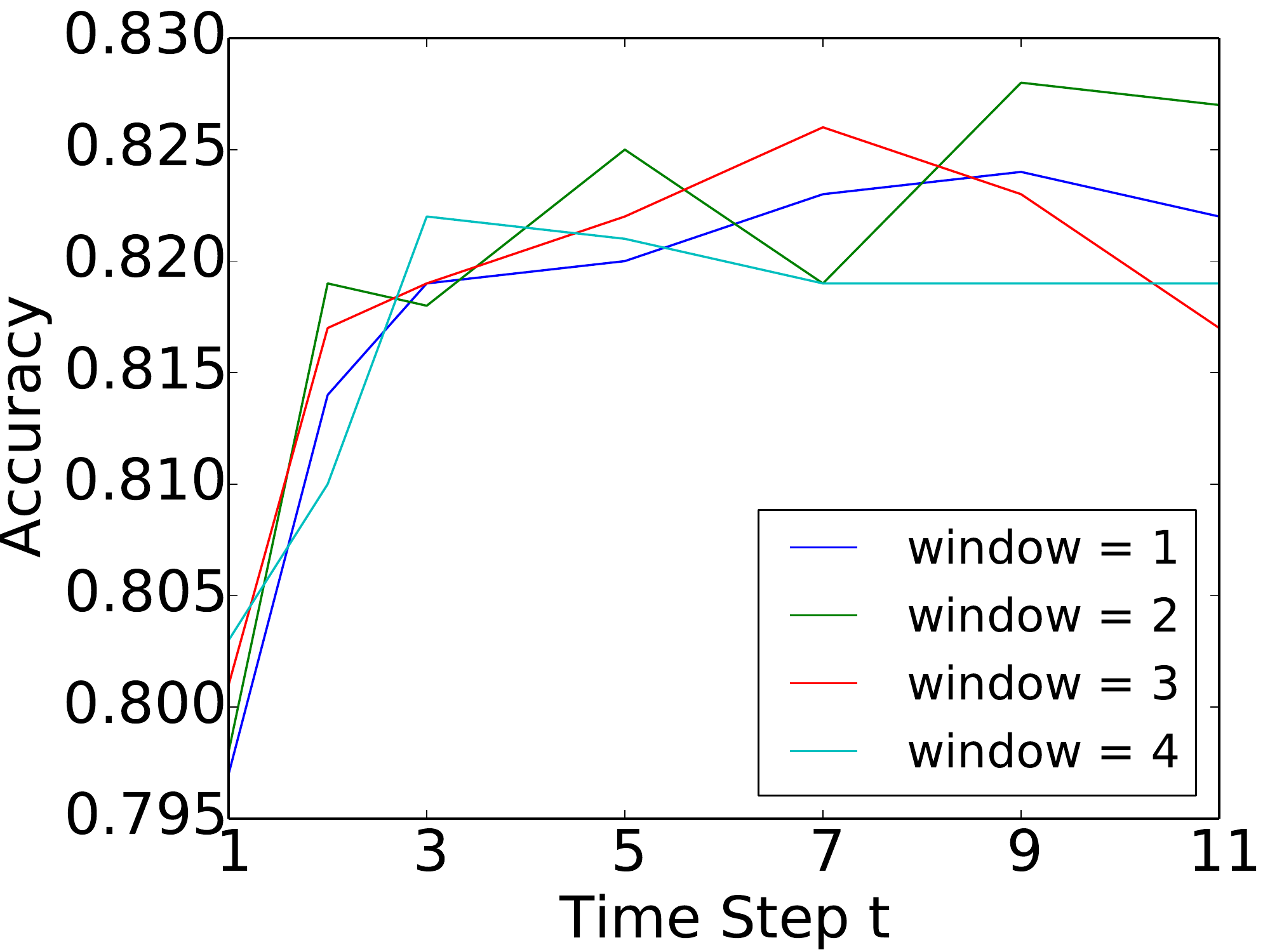}
\vspace{-0.8em}
\caption{Accuracies with various window sizes and time steps on movie review development set}
\label{fig:iteration_window}
\end{figure}

\subsection{Development Experiments}
We use the movie review development data to investigate different configurations of S-LSTMs and BiLSTMs. 
For S-LSTMs, the default configuration uses $\langle s \rangle$ and $\langle /s \rangle$ words for augmenting words of a sentence.
A hidden layer size of 300 and one sentence-level node are used.

\textbf{Hyperparameters}: Table \ref{tab:movie_dev} shows the development results of various S-LSTM settings, where Time refers to training time per epoch.
Without the sentence-level node, the accuracy of S-LSTM drops to 81.76\%, demonstrating the necessity of global information exchange. 
Adding one additional sentence-level node as described in Section~\ref{sec:m-lstm} does not lead to accuracy improvements, although the number of parameters and decoding time increase accordingly. 
As a result, we use only 1 sentence-level node for the remaining experiments.  
The accuracies of S-LSTM increases as the hidden layer size for each node increases from 100 to 300, but does not further increase when the size increases beyond 300. 
We fix the hidden size to 300 accordingly. 
Without using $\langle s \rangle$ and $\langle /s \rangle$, the performance of S-LSTM drops from 82.64\% to 82.36\%, showing the effectiveness of having these additional nodes. 
Hyperparameters for BiLSTM models are also set according to the development data, which we omit here.


{\bf State transition}. In Table \ref{tab:movie_dev}, the number of recurrent state transition steps of S-LSTM is decided according to the best development performance. 
Figure \ref{fig:iteration_window} draws the development accuracies of S-LSTMs with various window sizes against the number of recurrent steps. 
As can be seen from the figure, when the number of time steps increases from 1 to 11, the accuracies generally increase, before reaching a maximum value. 
This shows the effectiveness of recurrent information exchange in S-LSTM state transition.

On the other hand, no significant differences are observed on the peak accuracies given by different window sizes, although a larger window size (e.g. 4) generally results in faster plateauing. 
This can be be explained by the intuition that information exchange between distant nodes can be achieved using more recurrent steps under a smaller window size, as can be achieved using fewer steps under a larger window size.
Considering efficiency, we choose a window size of 1 for the remaining experiments, setting the number of recurrent steps to 9 according to Figure \ref{fig:iteration_window}.

\begin{table}[t]
	\centering
	\tabcolsep=0.1cm
	\begin{tabular}{|ccc|ccc|}
		\hline
		\multicolumn{3}{|c|}{\textbf{Model}}& \textbf{Time (s)} & \textbf{Acc} & \textbf{\# Param}\\ 
		\hline 	
		\multicolumn{3}{|c|}{LSTM}&67&80.72 &5,977K\\
		\multicolumn{3}{|c|}{BiLSTM}&106&81.73 &7,059K\\ 
		\multicolumn{3}{|c|}{2 stacked BiLSTM}&207&81.97&9,221K\\ 
		\multicolumn{3}{|c|}{3 stacked BiLSTM}&310&81.53 &11,383K\\
		\multicolumn{3}{|c|}{4 stacked BiLSTM}&411&81.37 &13,546K\\ 
		\multicolumn{3}{|c|}{S-LSTM}&65&82.64* &8,768K\\         
		\hline
		\multicolumn{3}{|c|}{CNN}&34&80.35&5,637K\\ 
        \multicolumn{3}{|c|}{2 stacked CNN}&40&80.97 &5,717K\\  
        \multicolumn{3}{|c|}{3 stacked CNN}&47&81.46 &5,808K\\  
        \multicolumn{3}{|c|}{4 stacked CNN}&51&81.39 &5,855K\\  
        \hline 	
		\multicolumn{3}{|c|}{Transformer (N=6)}&138&81.03&7,234K\\ 
        \multicolumn{3}{|c|}{Transformer (N=8)}&174&81.86 &7,615K\\  
        \multicolumn{3}{|c|}{Transformer (N=10)}&214&81.63 &8,004K\\  
		\hline 	
		\multicolumn{3}{|c|}{BiLSTM+Attention}&126&82.37 &7,419K\\
		\multicolumn{3}{|c|}{S-LSTM+Attention}&87&83.07* &8,858K\\
		\hline
	\end{tabular}
	\caption{\label{tab:movie_dev_2}Movie review development results}
\end{table}

\textbf{S-LSTM vs BiLSTM}: As shown in Table \ref{tab:movie_dev_2}, BiLSTM gives significantly better accuracies compared to uni-directional LSTM\footnote{$p<0.01$ using t-test. For the remaining of this paper, we use the same measure for statistical significance.}, with the training time per epoch growing from 67 seconds to 106 seconds. 
Stacking 2 layers of BiLSTM gives further improvements to development results, with a larger time of 207 seconds. 
3 layers of stacked BiLSTM does not further improve the results.
In contrast, S-LSTM gives a development result of 82.64\%, which is significantly better compared to 2-layer stacked BiLSTM, with a smaller number of model parameters and a shorter time of 65 seconds.

We additionally make comparisons with stacked CNNs and hierarchical attention \cite{vaswani2017attention}, shown in Table \ref{tab:movie_dev_2} (the CNN and Transformer rows), where $N$ indicates the number of attention layers.
CNN is the most efficient among all models compared, with the smallest model size. 
On the other hand, a 3-layer stacked CNN gives an accuracy of 81.46\%, which is also the lowest compared with BiLSTM,  hierarchical attention and S-LSTM.
The best performance of hierarchical attention is between single-layer and two-layer BiLSTMs in terms of both accuracy and efficiency.
S-LSTM gives significantly better accuracies compared with both CNN and hierarchical attention.

{\bf Influence of external attention mechanism}. 
Table \ref{tab:movie_dev_2} additionally shows the results of BiLSTM and S-LSTM when external attention is used as described in Section~\ref{sec:ext_attn}. 
Attention leads to improved accuracies for both BiLSTM and S-LSTM in classification, with S-LSTM still outperforming BiLSTM significantly. 
The result suggests that external techniques such as attention can play orthogonal roles compared with internal recurrent structures, therefore benefiting both BiLSTMs and S-LSTMs. 
Similar observations are found using external CRF layers for sequence labelling.

\begin{table}[t] 
	\centering
	\tabcolsep=0.03cm
	\begin{tabular}{ccccc|c|c|}
		\hline
		\multicolumn{4}{|l|}{\textbf{Model}}& \textbf{Accuracy} & \textbf{Train (s)}&\textbf{Test (s)}\\
		\hline
		\multicolumn{4}{|l|}{\,\,\newcite{socher2011semi}} &77.70& --& --\\
		\multicolumn{4}{|l|}{\,\,\newcite{socher2012semantic}} &79.00& --& --\\
		\multicolumn{4}{|l|}{\,\,\newcite{kim2014convolutional}} &81.50& --& --\\
        \multicolumn{4}{|l|}{\,\,\newcite{qian2016linguistically}} &81.50& --& --\\
		\hline
		\multicolumn{4}{|l|}{\,\,BiLSTM}&81.61&51&1.62\\
		\multicolumn{4}{|l|}{\,\,2 stacked BiLSTM}&81.94&98&3.18\\
		\multicolumn{4}{|l|}{\,\,3 stacked BiLSTM}&81.71&137&4.67\\
        \multicolumn{4}{|l|}{\,\,3 stacked CNN}&81.59&31&1.04\\
        \multicolumn{4}{|l|}{\,\,Transformer (N=8)}&81.97&89&2.75\\
		\multicolumn{4}{|l|}{\,\,S-LSTM}&\textbf{82.45}* & 41&1.53\\ 
		\hline
	\end{tabular}
\caption{\label{movie_review_classification}Test set results on movie review dataset (* denotes significance in all tables).}
\end{table}

\begin{table*}[t]
	\centering
	\begin{tabular}{cccc|c|c|c|c|c|c|}
		\hline
		\multicolumn{4}{|c|}{\textbf{Dataset}}& \textbf{SLSTM}&\textbf{Time (s)}&\textbf{BiLSTM} &\textbf{Time (s)}&\textbf{2 BiLSTM} &\textbf{Time (s)}\\
		\hline
		\multicolumn{4}{|c|}{Camera}&\textbf{90.02}*&50 (2.85)&87.05&115 (8.37)&88.07&221 (16.1)\\
		\multicolumn{4}{|c|}{Video}&\textbf{86.75}*&55 (3.95)&84.73&140 (12.59)&85.23&268 (25.86)\\
		\multicolumn{4}{|c|}{Health}&\textbf{86.5}&37 (2.17)&85.52&118 (6.38)&85.89&227 (11.16)\\
		\multicolumn{4}{|c|}{Music}&\textbf{82.04}*&52 (3.44)&78.74&185 (12.27)&80.45&268 (23.46)\\
		\multicolumn{4}{|c|}{Kitchen}&\textbf{84.54}*&40 (2.50)&82.22&118 (10.18)&83.77&225 (19.77)\\
		\multicolumn{4}{|c|}{DVD}&\textbf{85.52}*&63 (5.29)&83.71&166 (15.42)&84.77&217 (28.31)\\
		\multicolumn{4}{|c|}{Toys}&85.25&39 (2.42)&85.72&119 (7.58)&\textbf{85.82}&231 (14.83)\\
		\multicolumn{4}{|c|}{Baby}&\textbf{86.25}*&40 (2.63)&84.51&125 (8.50)&85.45&238 (17.73)\\
		\multicolumn{4}{|c|}{Books}&\textbf{83.44}*&64 (3.64)&82.12&240 (13.59)&82.77&458 (28.82)\\
		\multicolumn{4}{|c|}{IMDB}&\textbf{87.15}*&67 (3.69)&86.02&248 (13.33)&86.55&486 (26.22)\\
		\multicolumn{4}{|c|}{MR}&\textbf{76.2}&27 (1.25)&75.73&39 (2.27)&75.98&72 (4.63)\\
		\multicolumn{4}{|c|}{Appeal}&85.75&35 (2.83)&86.05&119 (11.98)&\textbf{86.35}*&229 (22.76)\\
		\multicolumn{4}{|c|}{Magazines}&\textbf{93.75}*&51 (2.93)&92.52&214 (11.06)&92.89&417 (22.77)\\
		\multicolumn{4}{|c|}{Electronics}&\textbf{83.25}*&47 (2.55)&82.51&195 (10.14)&82.33&356 (19.77)\\
		\multicolumn{4}{|c|}{Sports}&\textbf{85.75}*&44 (2.64)&84.04&172 (8.64)&84.78&328 (16.34)\\
		\multicolumn{4}{|c|}{Software}&\textbf{87.75}*&54 (2.98)&86.73&245 (12.38)&86.97&459 (24.68)\\
		\hline
		\multicolumn{4}{|c|}{\textbf{Average}} &\textbf{85.38}*&47.30 (2.98)&84.01&153.48 (10.29)&84.64&282.24 (20.2)\\
		\hline
	\end{tabular}
	\caption{\label{classification}Results on the 16 datasets of \newcite{liu2017adversarial}. Time format: train (test)}
\end{table*}

\begin{figure}[t]
\vspace{-0.6em}
\begin{center}	
\hspace{-.0em}
\subfigure[CoNLL03]{\includegraphics[width=0.9\linewidth]{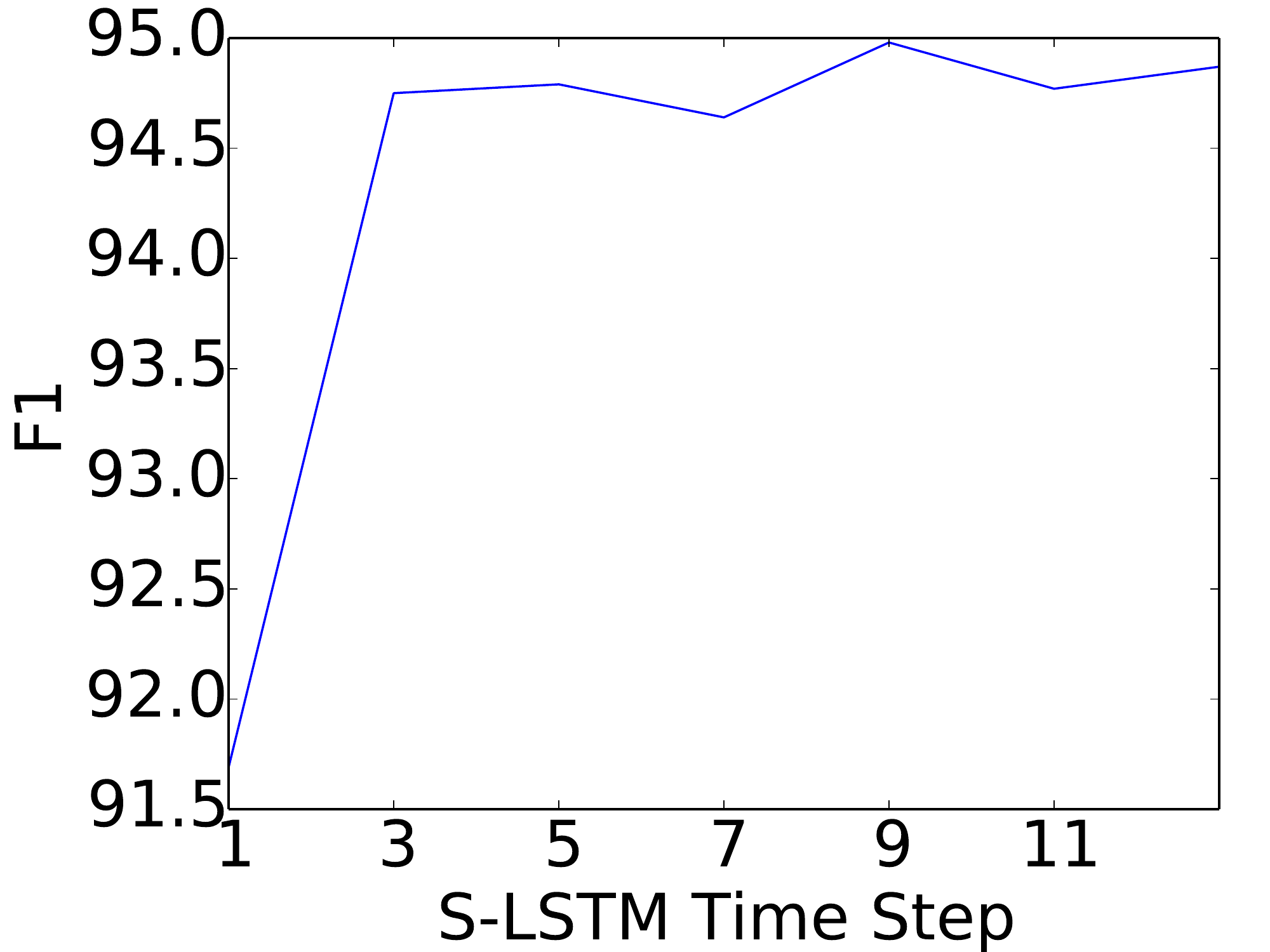}}\hskip 0.1pt 
\subfigure[WSJ]{\includegraphics[width=0.9\linewidth]{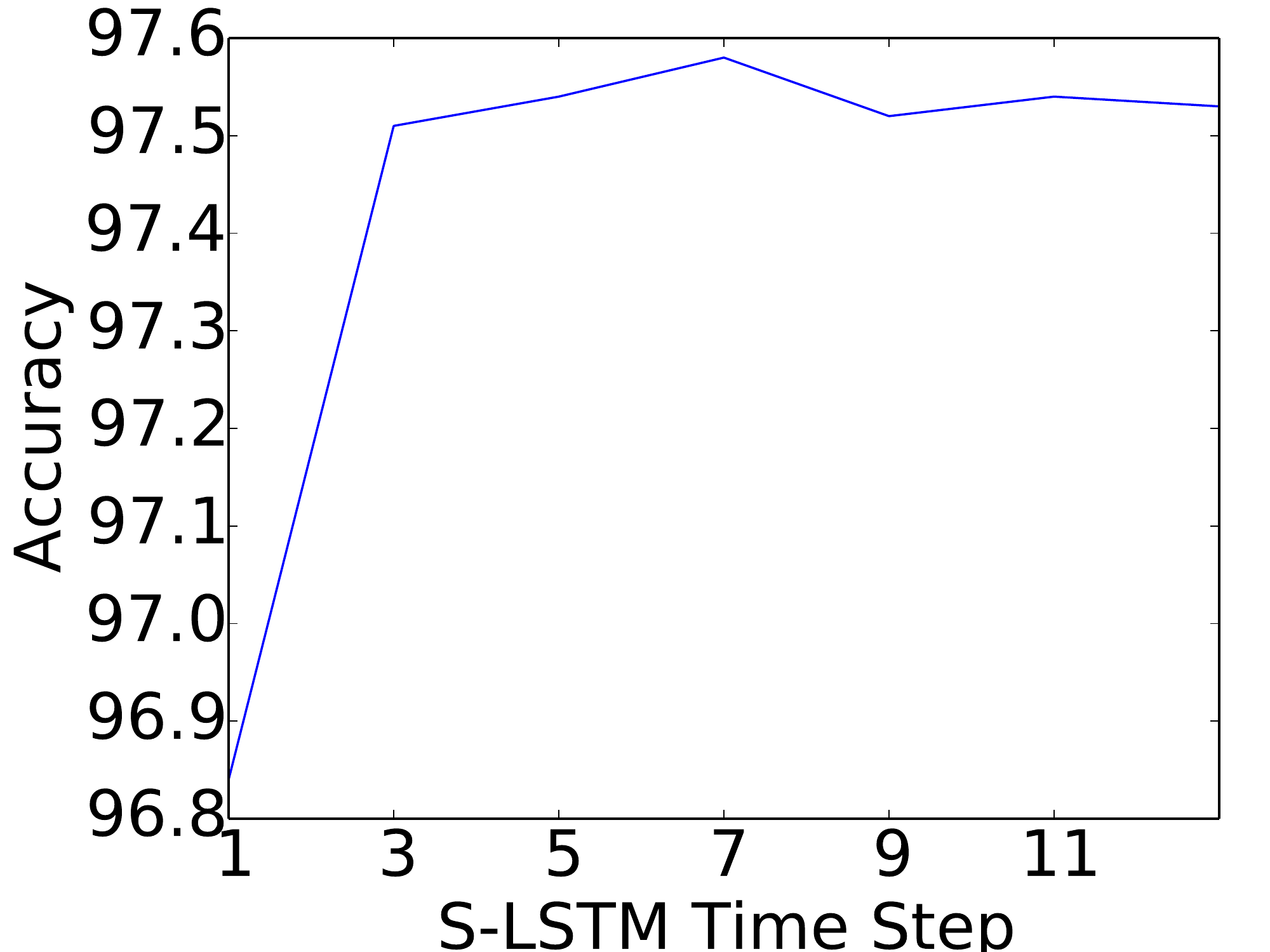}} \hskip 1pt
\end{center}
\vspace{-1em}
\caption{\label{sequence_dev}Sequence labelling development results.}
\end{figure}

\begin{table}[t] 
\centering
\tabcolsep=0.03cm
\begin{tabular}{|l|c|c|c|}
\hline
\textbf{Model} & \textbf{Accuracy}&\textbf{Train (s)} &\textbf{Test (s)}\\
\hline
\newcite{manning2011part} & 97.28&--&--\\
\newcite{collobert2011natural} & 97.29&--&--\\
\newcite{sun2014structure} & 97.36&--&--\\
\newcite{sogaard2011semisupervised} & 97.50 &--&--\\
\newcite{huang2015bidirectional} & \textbf{97.55} &--&--\\
\newcite{ma2016end} & \textbf{97.55} &--&--\\
\newcite{yang2017transfer} & \textbf{97.55} &--&--\\
\hline
BiLSTM&97.35&254&22.50\\ 
2 stacked BiLSTM&97.41&501&43.99\\ 
3 stacked BiLSTM&97.40&746&64.96\\ 
S-LSTM&\textbf{97.55}&237&22.16\\ 
\hline
\end{tabular}
\caption{\label{wsj_test}Results on PTB (POS tagging)}
\end{table}

\begin{table}[t]
\centering
\tabcolsep=0.03cm
\begin{tabular}{|l|c|c|c|}
\hline
\textbf{Model}& \textbf{F1} & \textbf{Train (s)} & \textbf{Test (s)} \\
\hline
\newcite{collobert2011natural} & 89.59 &--&--\\
\newcite{passos2014lexicon} &90.90&--&--\\
\newcite{luo2015joint} & 91.20&--&--\\
\newcite{huang2015bidirectional} & 90.10&--&--\\
\newcite{lample2016neural} & 90.94&--&--\\
\newcite{ma2016end} & 91.21&--&--\\
\hline
\newcite{yang2017transfer} & 91.26 & -- & -- \\
\newcite{rei:2017:Long} & 86.26 & -- & -- \\
\newcite{peters2017semi} & \textbf{91.93} & -- & -- \\
\hline
BiLSTM&90.96&82&9.89\\
2 stacked BiLSTM&91.02&159&18.88\\
3 stacked BiLSTM&91.06&235&30.97\\
S-LSTM & \textbf{91.57}*&79&9.78\\ 
\hline
\end{tabular}
\caption{Results on CoNLL03 (NER)}
\label{conll_test}
\vspace{-1.0em}
\end{table}

\subsection{Final Results for Classification}

The final results on the movie review and rich text classification datasets are shown in Tables \ref{movie_review_classification} and \ref{classification}, respectively. 
In addition to training time per epoch, test times are additionally reported.
We use the best settings on the movie review development dataset for both S-LSTMs and BiLSTMs. 
The step number for S-LSTMs is set to 9.

As shown in Table \ref{movie_review_classification}, the final results on the movie review dataset are consistent with the development results, where S-LSTM outperforms BiLSTM significantly, with a faster speed. 
Observations on CNN and hierarchical attention are consistent with the development results. 
S-LSTM also gives highly competitive results when compared with existing methods in the literature.

As shown in Table \ref{classification}, among the 16 datasets of \newcite{liu2017adversarial}, S-LSTM gives the best results on 12, compared with BiLSTM and 2 layered BiLSTM models. 
The average accuracy of S-LSTM is 85.6\%, significantly higher compared with 84.9\% by 2-layer stacked BiLSTM. 
3-layer stacked BiLSTM gives an average accuracy of 84.57\%, which is lower compared to a 2-layer stacked BiLSTM, with a training time per epoch of 423.6 seconds. 
The relative speed advantage of S-LSTM over BiLSTM is larger on the 16 datasets as compared to the movie review test test. 
This is because the average length of inputs is larger on the 16 datasets (see Section \ref{sec:analysis}).

\subsection{Final Results for Sequence Labelling}

Bi-directional RNN-CRF structures, and in particular BiLSTM-CRFs, have achieved the state of the art in the literature for sequence labelling tasks, including POS-tagging and NER. 
We compare S-LSTM-CRF with BiLSTM-CRF for sequence labelling, using the same settings as decided on the movie review development experiments for both BiLSTMs and S-LSTMs. 
For the latter, we decide the number of recurrent steps on the respective development sets for sequence labelling. 
The POS accuracies and NER F1-scores against the number of recurrent steps are shown in Figure \ref{sequence_dev} (a) and (b), respectively. 
For POS tagging, the best step number is set to 7, with a development accuracy of 97.58\%. 
For NER, the step number is set to 9, with a development F1-score of 94.98\%.

As can be seen in Table \ref{wsj_test}, S-LSTM gives significantly better results compared with BiLSTM on the WSJ dataset. 
It also gives competitive accuracies as compared with existing methods in the literature. 
Stacking two layers of BiLSTMs leads to improved results compared to one-layer BiLSTM, but the accuracy does not further improve with three layers of stacked LSTMs.

For NER (Table \ref{conll_test}), S-LSTM gives an F1-score of 91.57\% on the CoNLL test set, which is significantly better compared with BiLSTMs.
Stacking more layers of BiLSTMs leads to slightly better F1-scores compared with a single-layer BiLSTM.
Our BiLSTM results are comparable to the results reported by \newcite{ma2016end} and \newcite{lample2016neural}, who also use bidirectional RNN-CRF structures.
In contrast, S-LSTM gives the best reported results under the same settings.

In the second section of Table \ref{conll_test}, \newcite{yang2017transfer} use cross-domain data, obtaining an F-score of 91.26\%; \newcite{rei:2017:Long} perform multi-task learning using additional language model objectives, obtaining an F-score of 86.26\%; \newcite{peters2017semi} leverage character-level language models, obtaining an F-score of 91.93\%, which is the current best result on the dataset. 
All the three models are based on BiLSTM-CRF. 
On the other hand, these semi-supervised learning techniques are orthogonal to our work, and can potentially be used for S-LSTM also.

\subsection{Analysis}
\label{sec:analysis}

\begin{figure}[t]
\vspace{-0.6em}
\begin{center}	
\hspace{-.0em}
\subfigure[Movie review]{\includegraphics[width=0.9\linewidth]{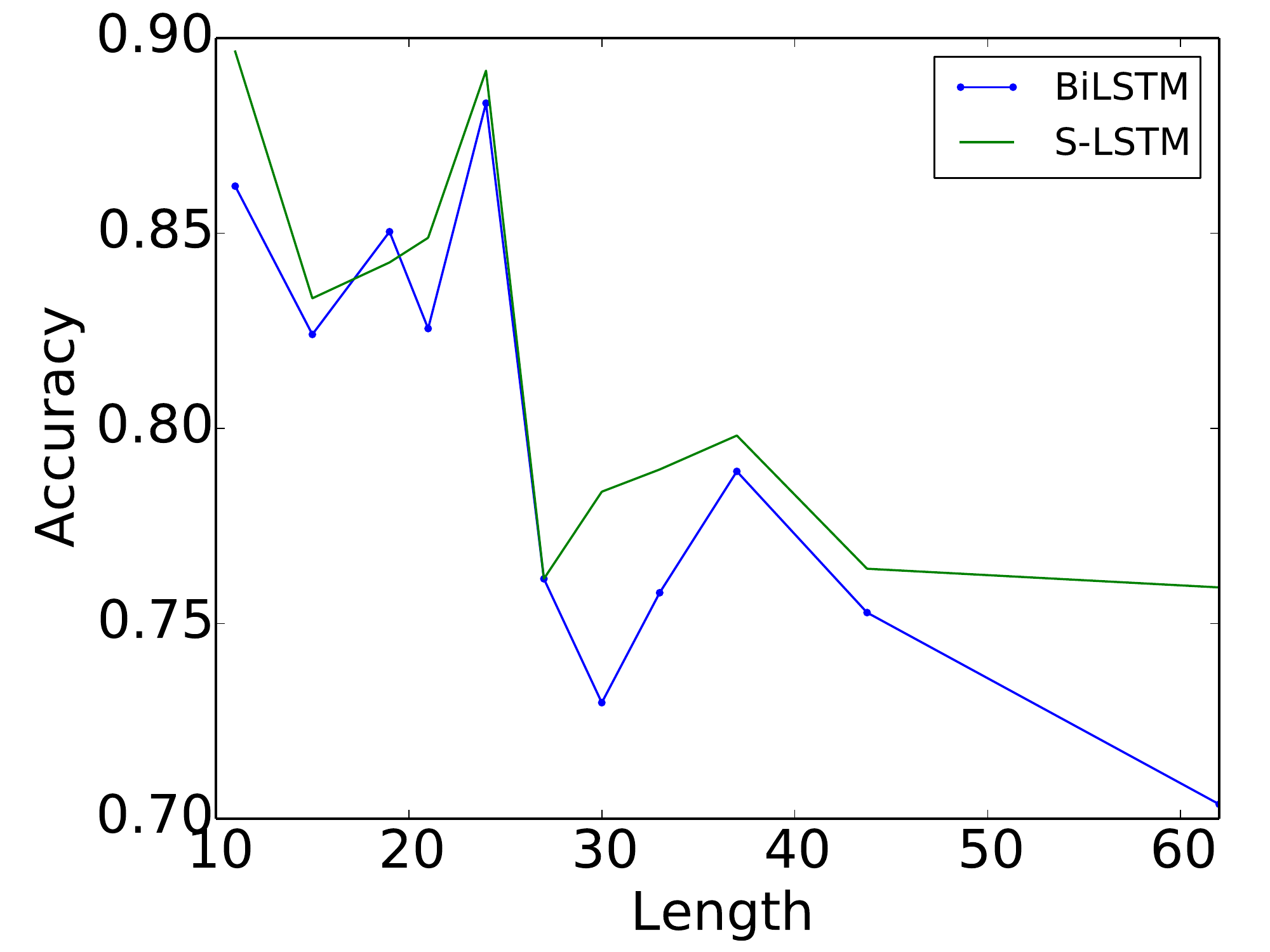}}\hskip 0.1pt 
\subfigure[CoNLL03]{\includegraphics[width=0.9\linewidth]{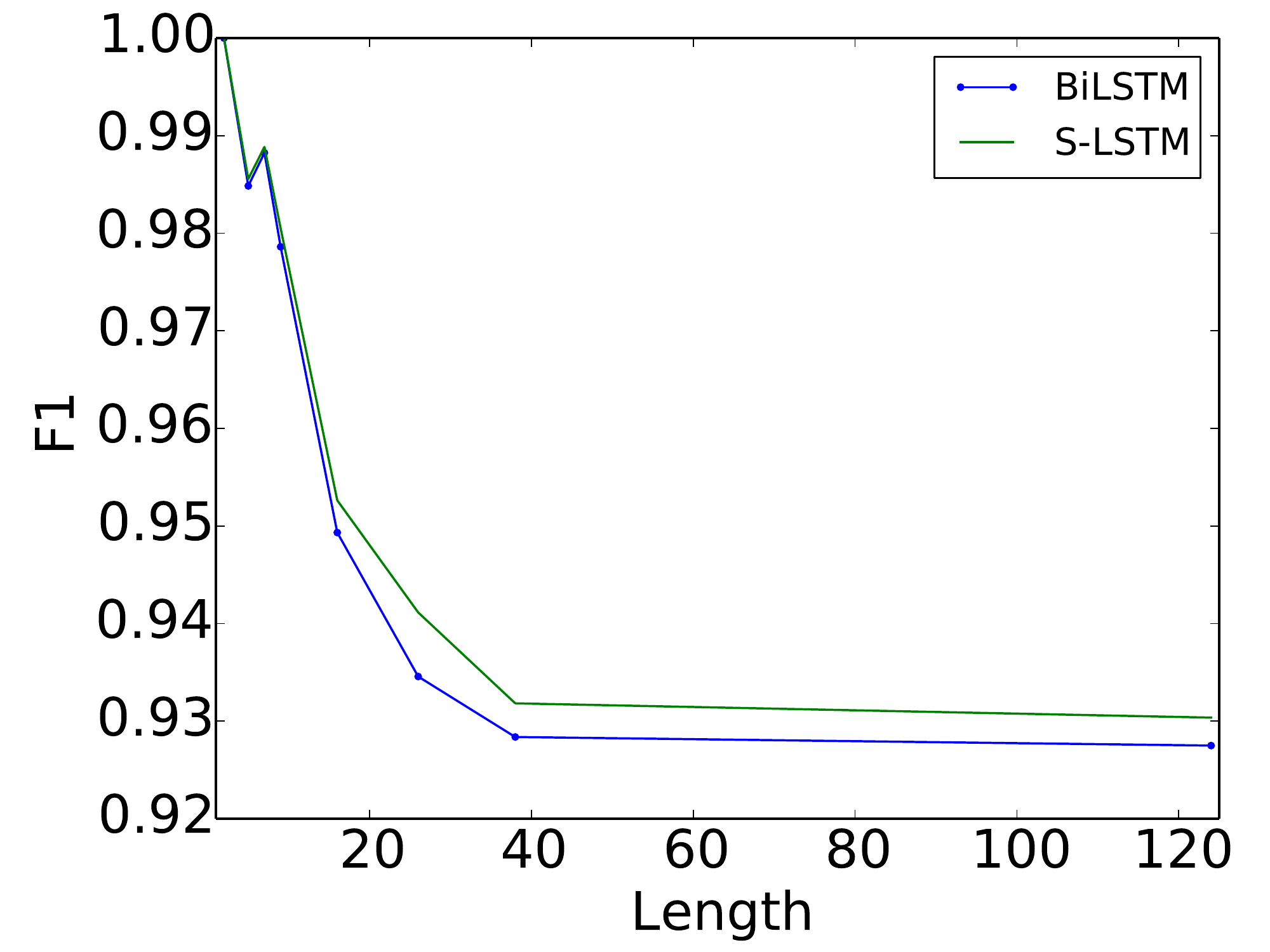}} \hskip 1pt
\end{center}
\vspace{-1em}
\caption{\label{fig:length_analysis}Accuracies against sentence length.}
\end{figure}

\begin{figure}[t]
\begin{center}
\includegraphics[width=0.9\linewidth]{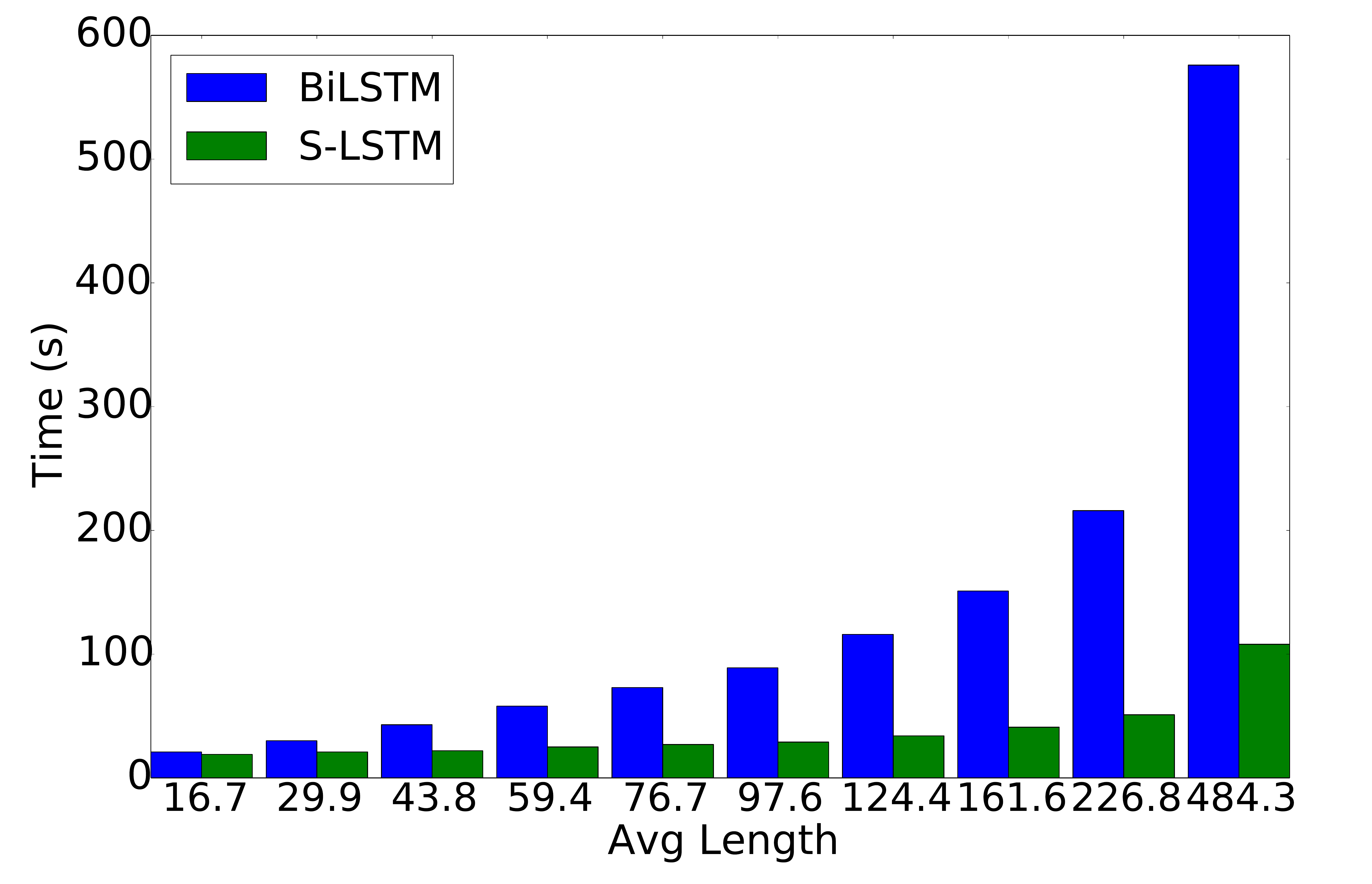}
\end{center}
\vspace{-1.5em}
\caption{\label{fig:length_time}Time against sentence length.}
\end{figure}

Figure \ref{fig:length_analysis} (a) and (b) show the accuracies against the sentence length on the movie review and CoNLL datasets, respectively, where test samples are binned in batches of 80. 
We find that the performances of both S-LSTM and BiLSTM decrease as the sentence length increases. 
On the other hand, S-LSTM demonstrates relatively better robustness compared to BiLSTMs.
This confirms our intuition that a sentence-level node can facilitate better non-local communication.

Figure \ref{fig:length_time} shows the training time per epoch of S-LSTM and BiLSTM on sentences with different lengths on the 16 classification datasets. 
To make these comparisons, we mix all training instances, order them by the size, and put them into 10 equal groups, the medium sentence lengths of which are shown.
As can be seen from the figure, the speed advantage of S-LSTM is larger when the size of the input text increases, thanks to a fixed number of recurrent steps. 

Similar to hierarchical attention \citep{vaswani2017attention}, there is a relative disadvantage of S-LSTM in comparison with BiLSTM, which is that the memory consumption is relatively larger.
For example, over the movie review development set, the actual GPU memory consumption by S-LSTM, BiLSTM, 2-layer stacked BiLSTM and 4-layer stacked BiLSTM are 252M, 89M, 146M and 253M, respectively.
This is due to the fact that computation is performed in parallel by S-LSTM and hierarchical attention.

\section{Conclusion}
We have investigated S-LSTM, a recurrent neural network for encoding sentences, which offers richer contextual information exchange with more parallelism compared to BiLSTMs. 
Results on a range of classification and sequence labelling tasks show that S-LSTM outperforms BiLSTMs using the same number of parameters, demonstrating that S-LSTM can be a useful addition to the neural toolbox for encoding sentences. 

The structural nature in S-LSTM states allows straightforward extension to tree structures, resulting in highly parallelisable tree LSTMs.
We leave such investigation to future work.
Next directions also include the investigation of S-LSTM to more NLP tasks, such as machine translation.

\section*{Acknowledge}
We thank the anonymous reviewers for their constructive and thoughtful comments.

\bibliography{acl2018}
\bibliographystyle{acl_natbib}

\end{document}